# An attempt to generate new bridge types from latent space of variational autoencoder

Hongjun Zhang

Wanshi Antecedence Digital Intelligence Traffic Technology Co., Ltd, Nanjing, 210016, China

583304953@QQ.com

**Abstract:** Try to generate new bridge types using generative artificial intelligence technology. The grayscale images of the bridge facade with the change of component width was rendered by 3dsMax animation software, and then the OpenCV module performed an appropriate amount of geometric transformation (rotation, horizontal scale, vertical scale) to obtain the image dataset of three-span beam bridge, arch bridge, cable-stayed bridge and suspension bridge. Based on Python programming language, TensorFlow and Keras deep learning platform framework, variational autoencoder was constructed and trained, and low-dimensional bridge-type latent space that is convenient for vector operations was obtained. Variational autoencoder can combine two bridge types on the basis of the original of human into one that is a new bridge type. Generative artificial intelligence technology can assist bridge designers in bridge-type innovation, and can be used as copilot.

## 1 Introduction

There are four types of bridge structures: beam bridge, arch bridge, cable-stayed bridge, and suspension bridge. Are these four bridge types all there are? Is there still a new bridge type that has not been discovered by humans? Can we find a way to assist human bridge designers in innovating bridge types? This article uses generative artificial intelligence technology to attempt to answer.

Since the birth of artificial intelligence technology in the middle of the last century, people have been trying to apply it to the field of bridge engineering. Expert System appeared in the field of bridge engineering in the last century [1], helping bridge engineers choose appropriate engineering techniques through human-machine question and answer methods. However, it declined due to the inherent differences among human expert groups on many technical issues and difficulties in maintaining expert system; Jiang Shaofei uses neural network for structural optimization and damage detection [2]; Shu Xin explored the application of artificial intelligence technology in bridge management and maintenance [3]; Sophia V. Kuhn used clustering, decision tree, and other algorithms to optimize the preliminary design of tied arch bridges [4]; Vera M. Balmer optimized the conceptual design of pedestrian overpasses using conditional variational autoencoder [5]. However, there is no research on the use of artificial intelligence technology for bridge type innovation at home and abroad.

Sampling from latent space to generate brand new images or editing existing images through vector operations in latent space is currently the most popular and successful creative artificial intelligence application. This article adopts the most basic and well-known variational autoencoder (VAE) architecture in the field to attempt bridge innovation (open source address of this article's dataset and source code: https://github.com/QQ583304953/Bridge-VAE ).

## 2 Introduction to latent space and variational autoencoder

### 2.1 Overview

Generative artificial intelligence technology has been applied in fields such as painting and literature, helping humans create, such as Stable Diffusion and ChatGPT. Its core algorithm architecture includes LSTM, DeepDream, Neural Style Transfer, VAE, GAN, Transformer, etc. This technology can learn the latent space of images, languages, etc., and then sample from the latent space to create new works with similar features as the works seen in the training data of the model

[6].

## 2.2 Latent space

Latent space is used to learn the inherent characteristics of data, simplify the expression of data (reduce its dimensions), and identify statistical patterns by highly summarizing the data [7]. Taking a plane projection dataset of an orthogonal straight bridge as an example (Figure 1):

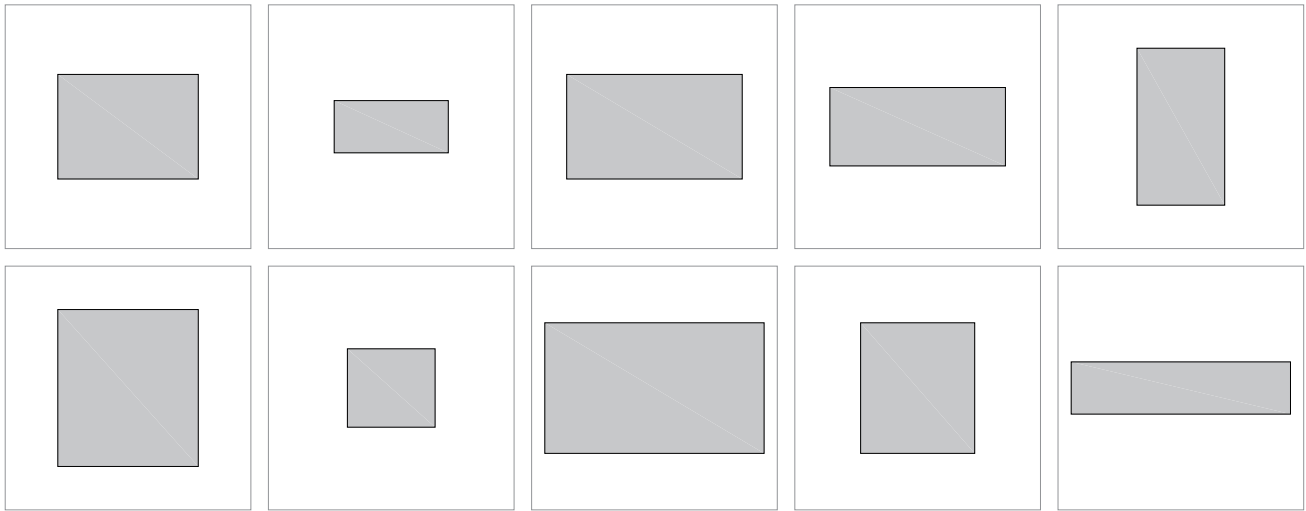

Fig.1 Plane projection dataset of bridge

For humans, there are clearly two features (corresponding to the two dimensions of the mathematical model), namely bridge length L and bridge width B, which can represent each sample in the dataset. Given the bridge length L and bridge width B, we can draw the corresponding plane projection of the bridge, even if it is not in a known dataset.

For machine learning, it first needs to determine that the bridge length L and bridge width B are the two latent space dimensions that best describe the dataset, and then learn the mapping function f. This function can map a point coordinate (bridge length L and bridge width B) in the latent space (vector space) into a bridge plane projection. The latent space of the bridge plane projection and the process of sampling from the latent space are shown in the figure (Figure 2):

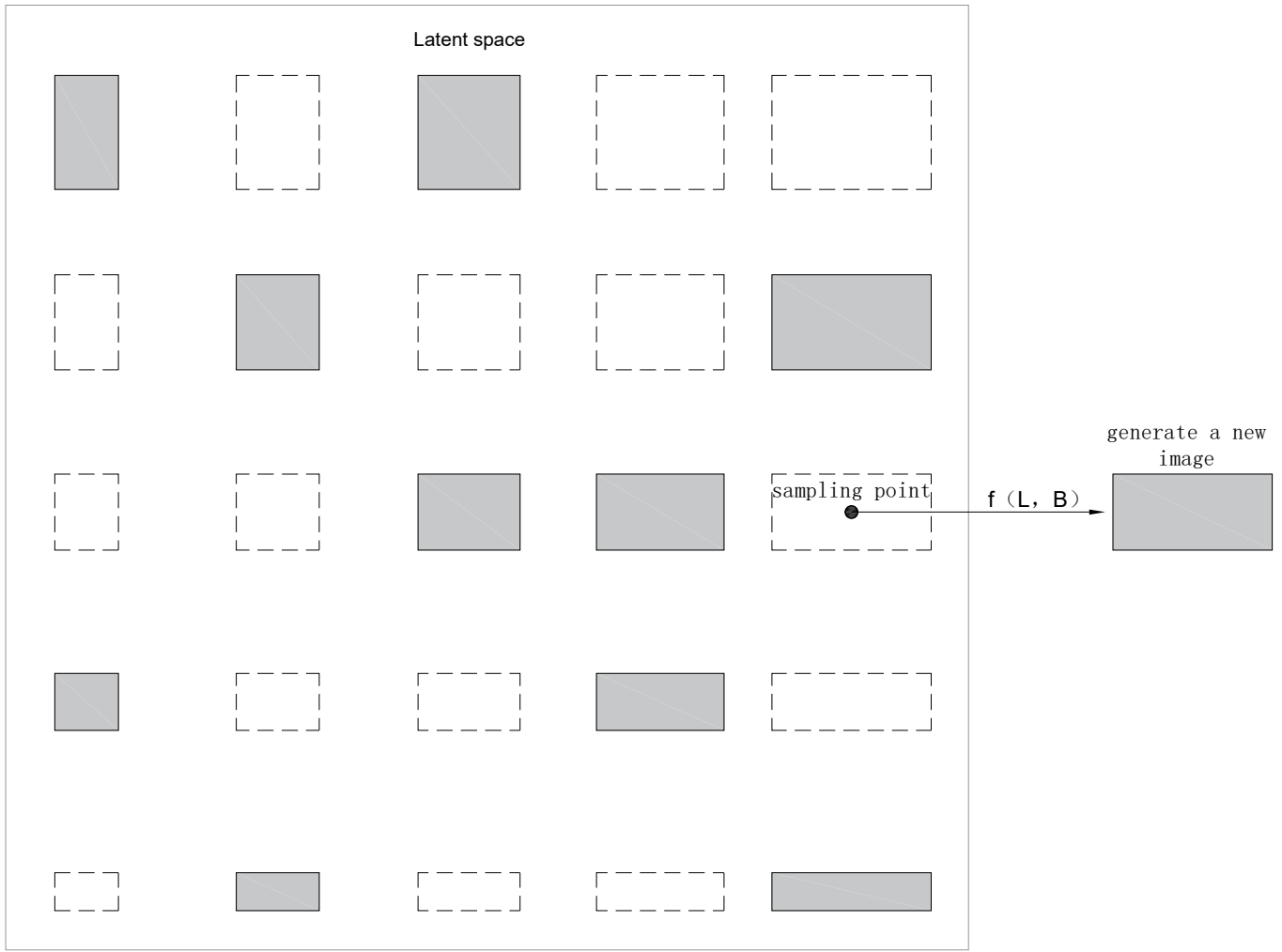

Fig.2 Schematic diagram of latent space and sampling from latent space

## 2.3 Variational autoencoder

Variational autoencoder has the advantages of low computational power consumption, easy training, continuous latent space structure, and easy vector operation. From the perspective of bridge-type innovation, it is the most suitable algorithm.

Let's briefly introduce AutoEncoder: it includes an encoder and a decoder. The framework diagram is shown in Figure 3:

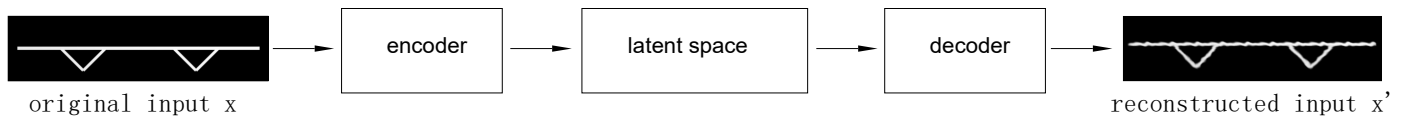

original input x                                                                                    reconstructed input x'

Fig.3 Architecture diagram of autoencoder

Variational autoencoder is to add appropriate randomness (noise) to the encoding result (representation of latent space) on the AutoEncoder. During training, the encoder of the variational autoencoder does not compress the original input x into a fixed encoding in the latent space, but instead converts the input into a statistical distribution (normal distribution, represented by mean and variance) in the latent space. Then, the decoder of the variational autoencoder uses the statistical distribution to randomly sample from the latent space, and decodes this random sampling into a reconstructed input x'[6].

Variational autoencoder is trained through two loss functions[6]: one is the reconstruction loss, which forces the decoded sample to be as identical as possible to the original input; The other is regularization loss, which is the difference between the statistical distribution of the sample in the latent space and the standard normal distribution. Because the encoding result is a statistical distribution, any two adjacent points in the latent space will be decoded as highly similar results, so there is continuity in the latent space. The low dimensional nature of latent space forces each dimension to represent a meaningful axis of change in the dataset, which makes it well structured and very suitable for operations through vector arithmetic.

## 3 An attempt to generate new bridge types from latent space of variational autoencoder

### 3.1 Self-built dataset

Due to cost constraints, it is difficult for the author to collect enough required real bridge photos. Therefore, 3dsmax animation software and opencv module were used to build a self-built dataset of bridge.

In order to reduce the difficulty of this attempt, each type of bridge is only self built in two subcategories (namely equal cross-section beam bridge, V-shaped pier rigid frame beam bridge, top-bearing arch bridge, bottom-bearing arch bridge, harp cable-stayed bridge, fan cable-stayed bridge, vertical_sling suspension bridge, and diagonal_sling suspension bridge), and all are three spans (beam bridge is 80+140+80m, while other bridge types are 67+166+67m).

The specific steps are as follows: ① Establish a bridge model in 3ds Max, animate the width change of the components (such as the beam height from 1m to 4m), and render 16 grayscale images of the bridge facade for each sub bridge type (Figure 4) in 512x128 pixels, png format; ② Subsequently, the OpenCV module was used to perform appropriate geometric transformations on the images, including rotation (with angles of -0.3, 0, and 0.3 degrees), horizontal scaling (with five interpolations ranging from 1 to 1.05), and vertical scaling (with five interpolations ranging from 1 to 1.1). In order to reduce training difficulty, translation geometric transformation was not used.

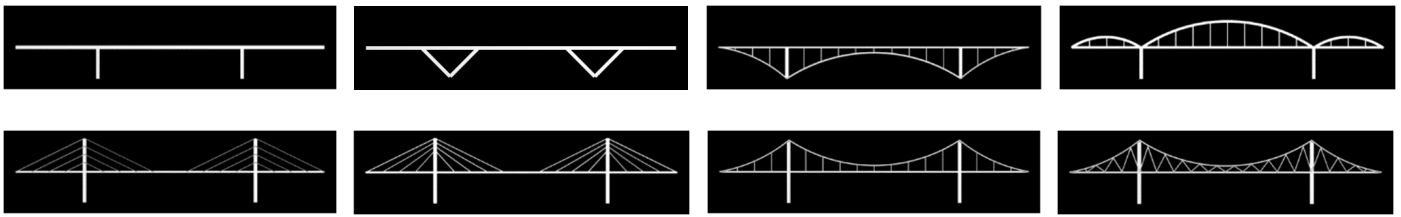

Fig.4 Grayscale image of each bridge facade

In the end, each sub bridge type obtained 16 * 3 * 5 * 5=1200 different images, resulting in a total of 9600 images in the entire dataset.

### 3.2 Construction of variational autoencoder

Based on the Python programming language, TensorFlow, and Keras deep learning platform framework, construct and train variational autoencoder. After testing, the 8-dimensional latent space can achieve relatively ideal results.

(1) Architecture of the encoder:

The job of the encoder is to obtain the input image and map it to an 8-dimensional normal distribution in the latent space.

First, create an input layer for the image, then passed to five Conv2D layers in sequence (activation function relu), and finally flatten and connect to two parallel Dense layers (linear transformation). The output results are the mean and variance (natural logarithm) of the normal distribution. The model summary is shown in the table below:

Tab.1 model summary of encoder

| Layer (type) | Output Shape | Param # | Connected to |
| --- | --- | --- | --- |
| input_1 (InputLayer) | (None, 128, 512, 1) | 0 | |
| conv2d (Conv2D) | (None, 64, 256, 64) | 640 | input_1[0][0] |
| batch_normalization | (None, 64, 256, 64) | 256 | conv2d[0][0] |
| activation (Activation) | (None, 64, 256, 64) | 0 | batch_normalization[0][0] |
| dropout (Dropout) | (None, 64, 256, 64) | 0 | activation[0][0] |
| conv2d_1 (Conv2D) | (None, 32, 128, 128) | 73856 | dropout[0][0] |
| batch_normalization_1 | (None, 32, 128, 128) | 512 | conv2d_1[0][0] |

| Layer (type) | Output Shape | Param # | Connected to |
|---|---|---|---|
| activation_1 (Activation) | (None, 32, 128, 128) | 0 | batch_normalization_1[0][0] |
| dropout_1 (Dropout) | (None, 32, 128, 128) | 0 | activation_1[0][0] |
| conv2d_2 (Conv2D) | (None, 16, 64, 128) | 147584 | dropout_1[0][0] |
| batch_normalization_2 | (None, 16, 64, 128) | 512 | conv2d_2[0][0] |
| activation_2 (Activation) | (None, 16, 64, 128) | 0 | batch_normalization_2[0][0] |
| dropout_2 (Dropout) | (None, 16, 64, 128) | 0 | activation_2[0][0] |
| conv2d_3 (Conv2D) | (None, 8, 32, 128) | 147584 | dropout_2[0][0] |
| batch_normalization_3 | (None, 8, 32, 128) | 512 | conv2d_3[0][0] |
| activation_3 (Activation) | (None, 8, 32, 128) | 0 | batch_normalization_3[0][0] |
| dropout_3 (Dropout) | (None, 8, 32, 128) | 0 | activation_3[0][0] |
| conv2d_4 (Conv2D) | (None, 4, 16, 128) | 147584 | dropout_3[0][0] |
| batch_normalization_4 | (None, 4, 16, 128) | 512 | conv2d_4[0][0] |
| activation_4 (Activation) | (None, 4, 16, 128) | 0 | batch_normalization_4[0][0] |
| dropout_4 (Dropout) | (None, 4, 16, 128) | 0 | activation_4[0][0] |
| flatten (Flatten) | (None, 8192) | 0 | dropout_4[0][0] |
| dense (Dense) | (None, 8) | 65544 | flatten[0][0] |
| dense_1 (Dense) | (None, 8) | 65544 | flatten[0][0] |
| Total params: 650640 | | | |
| Trainable params: 649488 | | | |
| Non-trainable params: 1152 | | | |

(2) Calculation of random sampling coordinates from latent space for samples during training

Each input sample is transformed into an 8-dimensional normal distribution in the latent space by the encoder. For the convenience of discussion, taking a certain one dimension as an example, calculate the random sampling coordinates of the sample in that dimension. According to the principles of mathematical statistics, the standardized value of a random variable = (original value -mean)/standard deviation, so the original value = mean +standard deviation * standardized value. The specific calculation formula is as follows:

$z = z\_mean + exp(0.5 * z\_log\_var) * epsilon$

In this formula: z is the original value of a random variable, where is the random sampling coordinates from the latent space; z_mean is the mean of a random variable, where it is the mean output by the encoder; z_log_var is the variance (natural logarithm) of a random variable, where it is the variance output by the encoder; epsilon is the standardized value of a random variable, where represents the randomness of sampling.

Calculation example: z_mean=1.72, z_log_var=-4.27, epsilon=3 (the standard normal distribution values fall in the range of [-3, 3] by 99.74%), z=1.72+ exp(-0.5*4.27) * 3=2.07.

(3) Architecture of the decoder:

The job the decoder is to map coordinate points in the latent space into images.

First, create a coordinate point input layer and a Dense layer, and then passed to five Conv2D Transfer layers in sequence (the first four activation functions relu and the last activation function sigmoid). The output result is the reconstructed image corresponding to the coordinate point. The model summary is shown in the table below:

Tab.2 model summary of decoder

| Layer (type) | Output Shape | Param # |
|---|---|---|
| input_2 (InputLayer) | (None, 8) | 0 |
| dense_2 (Dense) | (None, 8192) | 73728 |
| reshape (Reshape) | (None, 4, 16, 128) | 0 |
| conv2d_transpose | (None, 8, 32, 128) | 147584 |
| batch_normalization_5 | (None, 8, 32, 128) | 512 |
| activation_5 (Activation) | (None, 8, 32, 128) | 0 |
| dropout_5 (Dropout) | (None, 8, 32, 128) | 0 |
| conv2d_transpose_1 | (None, 16, 64, 128) | 147584 |
| batch_normalization_6 | (None, 16, 64, 128) | 512 |
| activation_6 (Activation) | (None, 16, 64, 128) | 0 |
| dropout_6 (Dropout) | (None, 16, 64, 128) | 0 |
| conv2d_transpose_2 | (None, 32, 128, 128) | 147584 |
| batch_normalization_7 | (None, 32, 128, 128) | 512 |
| activation_7 (Activation) | (None, 32, 128, 128) | 0 |
| dropout_7 (Dropout) | (None, 32, 128, 128) | 0 |
| conv2d_transpose_3 | (None, 64, 256, 64) | 73792 |
| batch_normalization_8 | (None, 64, 256, 64) | 256 |
| activation_8 (Activation) | (None, 64, 256, 64) | 0 |
| dropout_8 (Dropout) | (None, 64, 256, 64) | 0 |
| conv2d_transpose_4 | (None, 128, 512, 1) | 577 |
| Total params: 592641 | | |
| Trainable params: 591745 | | |
| Non-trainable params: 896 | | |

## 3.3 Loss Function

(1) Reconstruction loss

Using binary cross entropy to calculate the difference between input sample and decoded reconstructed image[8].

reconstruction_loss=1/n*∑ {y*ln(1/y′)+(1-y)*ln[1/(1-y′)]}

In this formula: reconstruction_loss is the reconstruction loss of a single sample; y is the pixel value of the original image; y′ is the pixel value of the decoded reconstructed image; n is the number of pixels; To avoid a denominator of 0, a small value (2e-07) needs to be added to the denominator, which is not expressed in the formula.

Calculation example: y=[0.0, 0.1, 0.9, 1.0], y′=[0.0, 0.9, 0.99, 0.1],
reconstruction_loss=1/4*(0+2.083+0.470+2.303)=1.214.

(2) Regularization loss

Using KL divergence (relative entropy) to calculate the difference between the statistical distribution of input sample in the latent space and the standard normal distribution[7].

kl_loss =1/m*∑ {- 0.5 *[1 + z_log_var - square(z_mean) - exp(z_log_var)]}

In this formula: kl_loss is the regularization loss of a single sample; z_mean is the average value output by the encoder; z_log_var is the variance output by the encoder (natural logarithm); m is the number of latent space dimensions.

Calculation example: z_mean=[4.5, 3.3], z_log_var=[-3.3, -3.7], kl_loss=1/2*(11.3+6.8)=9.1.

(3) The loss function of variational autoencoder is the sum of reconstruction loss and regularization loss, with multiple samples taking the average loss.

total_loss = reconstruction_loss + kl_loss* coefficient

In this formula: total_loss is the loss function of the variational autoencoder; coefficient is the relative weight coefficient of loss.

If the proportion of reconstruction loss is too large, the statistical distribution of the training sample encoding will deviate too far from the standard normal distribution, and the continuity of the latent space will be poor; If the proportion of regularization loss is too large, the decoded reconstructed image is not clear enough.

## 3.4 Training

Update neural network parameters using the RMSProp optimizer. Label dictionary of the dataset is: {'Arch Bottom_bear': 0, 'Arch Top_bear': 1, 'Beam Three_span': 2, 'Beam V_type': 3, 'Cable Fan_shaped': 4, 'Cable Harp_shaped': 5, 'Suspension Diagonal_sling': 6, 'Suspension Vertical_sling': 7}.

At the beginning of the training, the default initial value of the neural network parameters is very small random value, so the pixel values of the decoded image are all close to 0 (grayscale image 0 corresponds to black, 1 corresponds to white), reconstruction_ Loss=3000 * 15/(128 * 512)=0.7 or so (approximately 3000 pixels per input sample have values close to 1, others are 0), and as the neural network parameters update, the reconstruction loss will gradually decrease. Similarly, at the beginning of the training, the mean and variance are 0, so the regularization loss is 0. Then, as the neural network parameters are updated, the regularization loss rapidly increases to become a control element, and the optimizer suppresses it, gradually reducing it.

After debugging, it was found that the relative weight coefficient of loss of 0.001 is more suitable, the decoded image is clear, and the distribution of training samples in the latent space is reasonable. Due to the inability to display more than three dimensions simultaneously, one-dimensional histograms (Figure 5, where the smooth curve is the standard normal distribution curve for comparison) and two-dimensional scatter plots (Figure 6) are used to display the distribution of all training samples in the latent space.

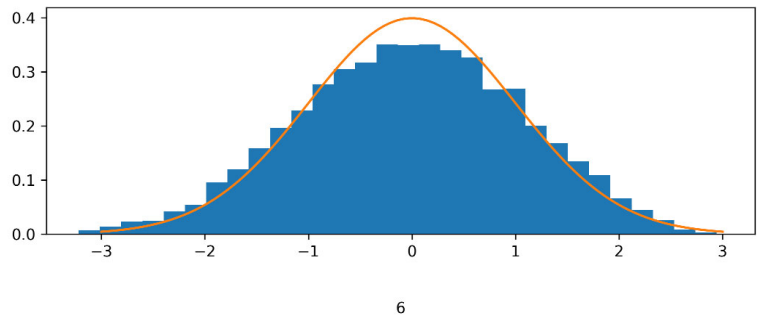

Fig.5 Frequency distribution histogram of all samples in the 6th dimension

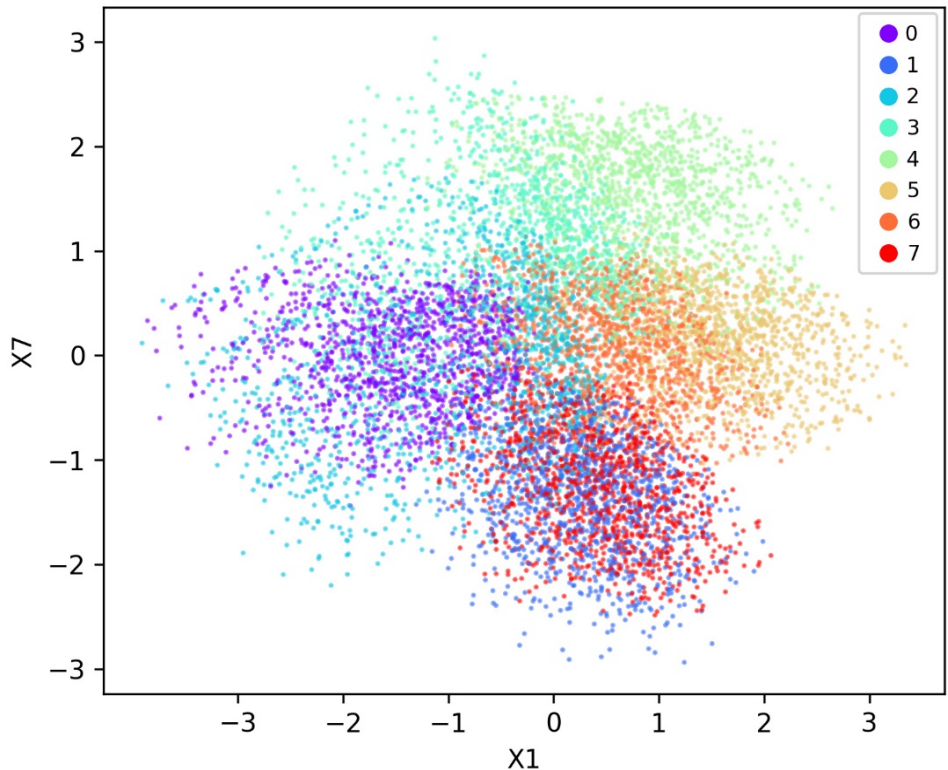

Fig.6 Scatter distribution map of all samples in the 1 and 7 dimensional planes

3.5 Exploring new bridge types through latent space sampling

To gain a detailed understanding of the latent space of 8 dimensions, theoretically, it is possible to sample from the latent space according to the set dense spatial grid. If 9 points are equidistant from -4 to+4 for each dimension, the total number of samples is approximately 40 million images (to the 8th power of 9), indicating that so many samples are difficult to operate in practice.

The sampling ideas for this practice are as follows:

(1) Sampling from the periphery of the training sample distribution boundary in latent space

Sampling is divided into 4 times, with only 2 points per dimension taken each time (coordinates are [-100,+100], [-5,+5], [-4,+4], [-3,+3]). The number of samples per time is 256 images (to the 8th power of 2). The results of a certain sampling are as follows (Figure 7):

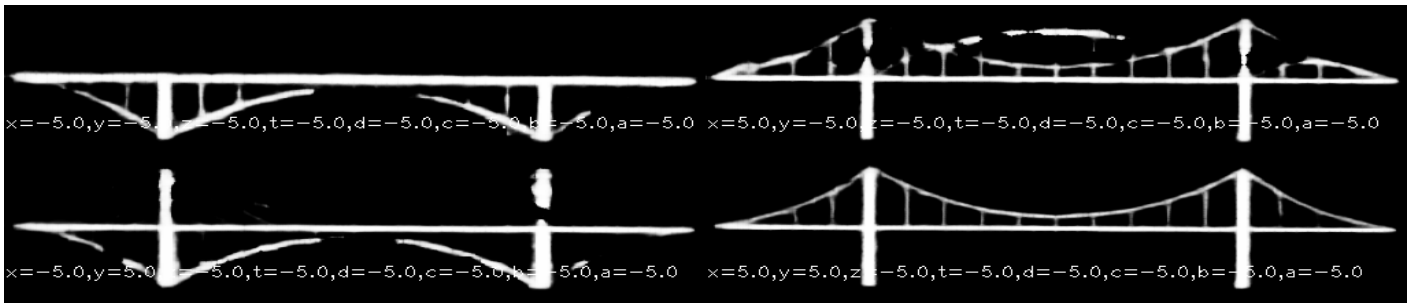

Fig.7 Bridge facade generated by sampling outside the sample distribution boundary

After manual exploration, it was not possible to generate original new bridge types, but there are several new combination bridge types that differ from the training samples. Due to the distance between the sampling points and the training samples, the image quality is low.

(2) Sampling from within the distribution boundary of training samples in latent space

Firstly, calculate the center coordinates of all samples for each sub bridge type, and then draw morphing Images between the two sub bridge types.

Choose any two of the eight sub bridge types, with a total of 28 combinations. After manual exploration, it was not possible to generate original new bridge types, but there are several new combination bridge types that differ from the training samples. Based on the thinking of engineering structure, five kinds of new composite bridge types are obtained (Figure 8).

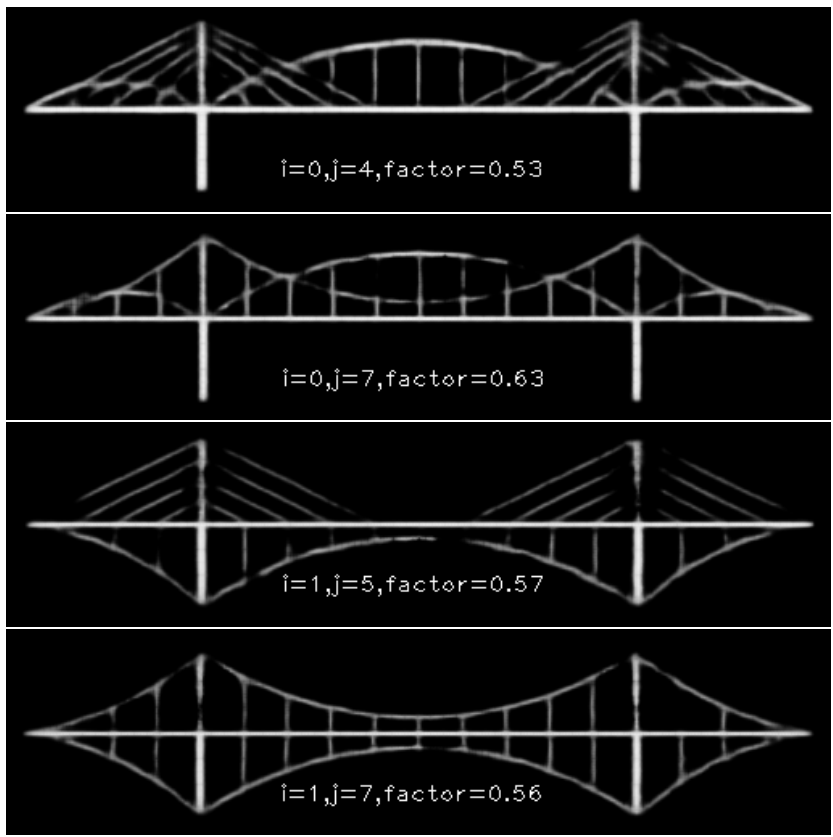

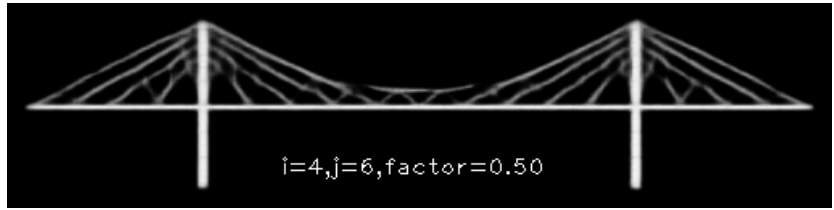
Fig.8 Five new combined bridge types with feasible technology

The above combination bridge types may not be suitable for super large span bridges due to their complex stress or unreasonable structure, but they can be used for urban landscape bridges. Some combination bridge types, such as hybrid cable-supported bridge system, have been conceptualized and practiced in the bridge industry [9].

# 4 Conclusion

Variational autoencoder can combine two bridge types on the basis of the original of human into one that is a new bridge type. Originality is an advantage of humans, while combinatorial creation is an advantage of artificial intelligence. Generative artificial intelligence has enormous potential and should have a place in the field of bridge design, serving as the copilot of bridge designers.

The data determines the upper limit of machine learning [10], and there are only eight sub bridge types in this dataset, which seriously restricts the innovative ability of variational autoencoder. If the size of the dataset is increased, and even three-dimensional bridges are used, then low-dimensional latent space will tend to use basic engineering structural units to describe training samples, making it easier to generate composite bridges (even generate new original bridge types). If a larger scale neural network model is adopted, it cannot be ruled out that it will generate emergent abilities similar to large models.